\documentclass[twoside,leqno,twocolumn]{article}  
\usepackage{ltexpprt} 

\usepackage{amsmath,amssymb,xspace,subfigure,paralist,wrapfig,colortbl}
\usepackage{graphicx}
\usepackage{xspace}
\usepackage{color,xspace}
\usepackage{euscript}
\usepackage{algorithm,algorithmic}
\newtheorem{defn}{Definition}[section]

\newcommand{\emd}{\ensuremath{d_T}\xspace}
\newcommand{\hausdorff}{\ensuremath{d_H}\xspace}
\renewcommand{\c}[1]{\ensuremath\EuScript{#1}}
\renewcommand{\b}[1]{\ensuremath\mathbb{#1}}
\newcommand{\eps}{\varepsilon}
\newcommand{\consensus}{\ensuremath{\c{P}_{\text{con}}}\xspace}
\newcommand{\reals}{\mathbb{R}}

\newcommand{\liftemd}{\textsc{LiftEMD}\xspace}
\newcommand{\liftcd}{\textsc{LiftKD}\xspace}
\newcommand{\lifthausdorff}{\textsc{LiftH}\xspace}
\newcommand{\liftkmeans}{\textsc{LiftKm}\xspace}
\newcommand{\lifthac}{\textsc{LiftHAC}\xspace}
\newcommand{\mallows}{\textsc{CC}\xspace}
\newcommand{\adco}{\ensuremath{D_{\textsc{ADCO}}}\xspace}
\newcommand{\cdistance}{\textsc{CDistance}\xspace}
\newcommand{\liftSSD}{\ensuremath{\textsc{liftSSD}}\xspace}
\newcommand{\etal}{\textit{et al}\xspace}

\numberwithin{equation}{section}

\newcommand{\denselist}{\itemsep -2pt\parsep=-1pt\partopsep -2pt}

\usepackage{color}
\definecolor{MyGray}{rgb}{0.9,0.9,0.9}
\definecolor{MyRed}{rgb}{1.0,0.0,0.0}
\definecolor{MyBlue}{rgb}{0.0,0.0,1.0}
\definecolor{MyGreen}{rgb}{0.1,0.7,0.1}
\makeatletter

\newenvironment{keywords}{
       \list{}{\advance\topsep by0.35cm\relax\small
       \leftmargin=1cm
       \labelwidth=0.35cm
       \listparindent=0.35cm
       \itemindent\listparindent
       \rightmargin\leftmargin}\item[\hskip\labelsep
                                     \bfseries Keywords:]}
     {\endlist}

\makeatother

\newlength{\ppicwd}

\begin{document}

\title{\Large Spatially-Aware Comparison and Consensus for Clusterings\thanks{This research was supported by NSF award CCF-0953066 and a subaward to the University of Utah under NSF award 0937060 to the Computing Research Association.}}
\author{Parasaran Raman
\and 
Jeff M. Phillips
\and
Suresh Venkatasubramanian
\date{School of Computing, University of Utah. \\ \textsl{\{praman,jeffp,suresh\}@cs.utah.edu}}
}

\maketitle

\begin{abstract} \small\baselineskip=9pt 
This paper proposes a new distance metric between clusterings that incorporates information about the spatial distribution of points and clusters. Our approach builds on the idea of a Hilbert space-based representation of clusters as a combination of the representations of their constituent points. We use this representation and the underlying metric to design a spatially-aware consensus clustering procedure. This consensus procedure is implemented via a novel reduction to Euclidean clustering, and is both simple and efficient. All of our results apply to both soft and hard clusterings. We accompany these algorithms with a detailed experimental evaluation that demonstrates the efficiency and quality of our techniques.
\end{abstract}

\vspace{-.2in}
\vspace{-.3in}

\section{Introduction}
\label{sec:intro}
The problem of \emph{metaclustering} has become important in recent years as researchers have tried to combine the strengths and weaknesses of different clustering algorithms to find patterns in data. A popular metaclustering problem is that of finding a consensus (or ensemble) partition\footnote{We use the term \emph{partition} instead of \emph{clustering} to represent a set of clusters decomposing a dataset.  This avoids confusion between the terms 'cluster', 'clustering' and the procedure used to \emph{compute} a partition, and will help us avoid phrases like, ``We compute consensus clusterings by clustering clusters in clusterings!"} from among a set of candidate partitions.  Ensemble-based clustering has been found to be very powerful when different clusters are connected in different ways, each detectable by different classes of clustering algorithms~\cite{nmi}.  For instance, no single clustering algorithm can detect clusters of symmetric Gaussian-like distributions of different density and clusters of long thinly-connected paths; but these clusters can be correctly identified by combining multiple techniques (i.e. $k$-means and single-link)~\cite{survey}.

Other related and important metaclustering problems include finding a different and yet informative partition to a given one, or finding a set of partitions that are mutually diverse (and therefore informative). 
In all these problems, the key underlying step is comparing two partitions and quantifying the difference between them. Numerous metrics (and similarity measures) have been proposed to compare partitions, and for the most part they are based on comparing the combinatorial structure of the partitions. This is done either by examining pairs of points that are grouped together in one partition and  separated in another~\cite{Ran71,jaccard,mirkin,fm}, or by information-theoretic considerations stemming from building a histogram of cluster sizes and normalizing it to form a distribution~\cite{vi,nmi}. 

\begin{figure}[b!!!]
\vspace{-.25in}
\begin{center}
\subfigure[Reference Partition (\textit{RP}) \label{fig:spatial-a}]{\includegraphics[angle=90,width=.3\linewidth]{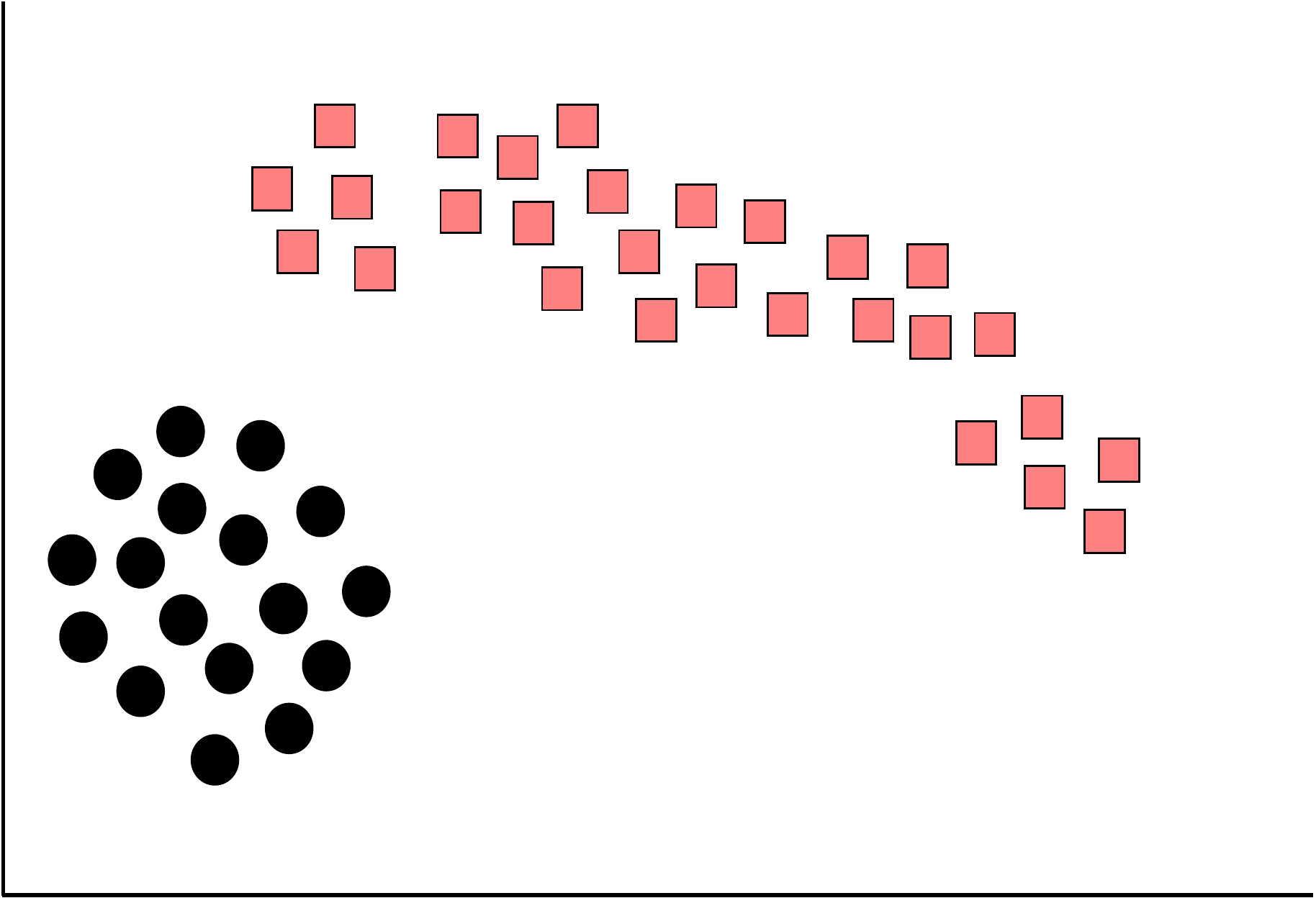}}
\vspace{-.2in}
\subfigure[First Partition (\textit{FP})\label{fig:spatial-b}]{\includegraphics[angle=90,width=.3\linewidth]{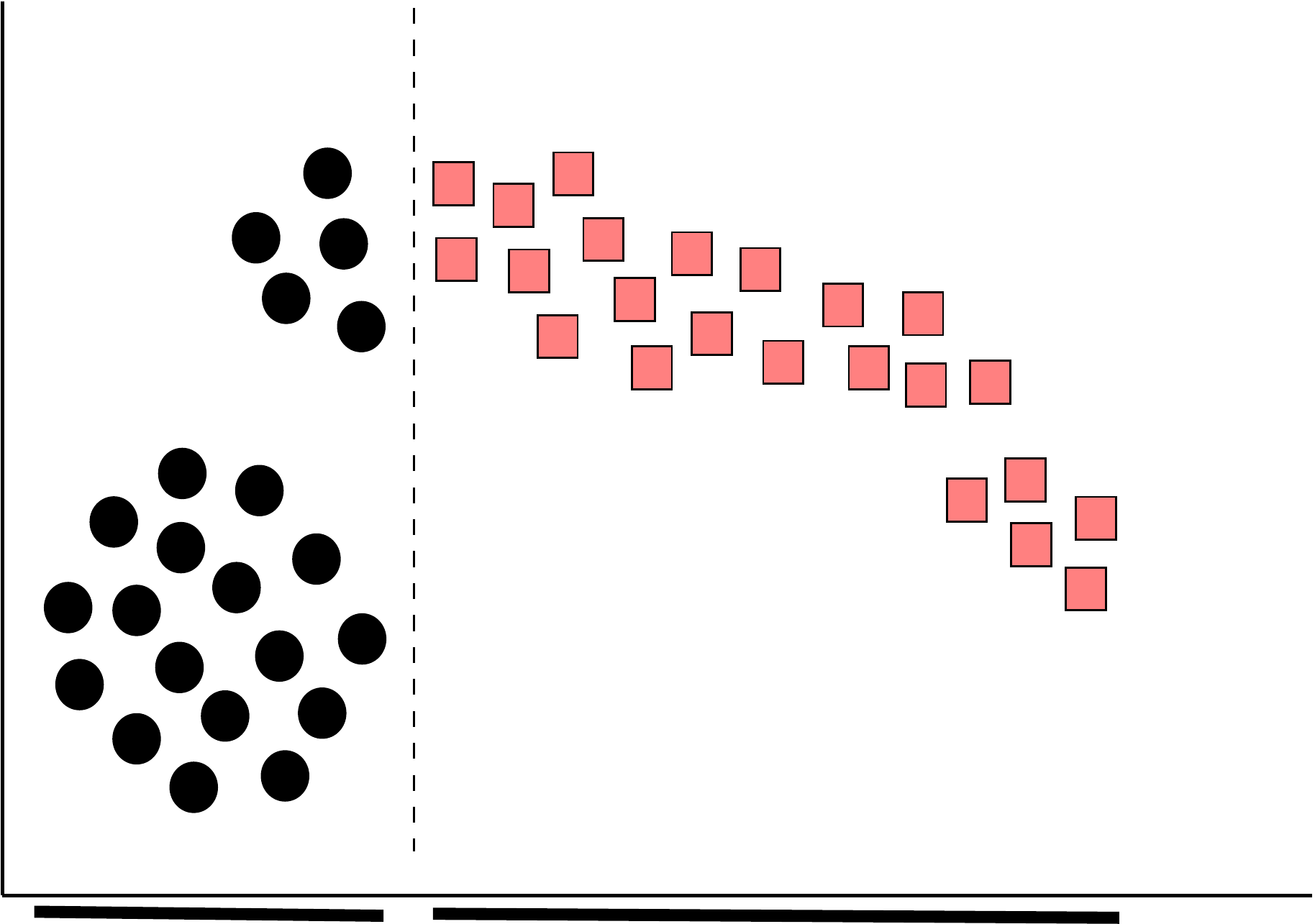}}
\subfigure[Second Partition (\textit{SP})\label{fig:spatial-c}]{\includegraphics[angle=90,width=.3\linewidth]{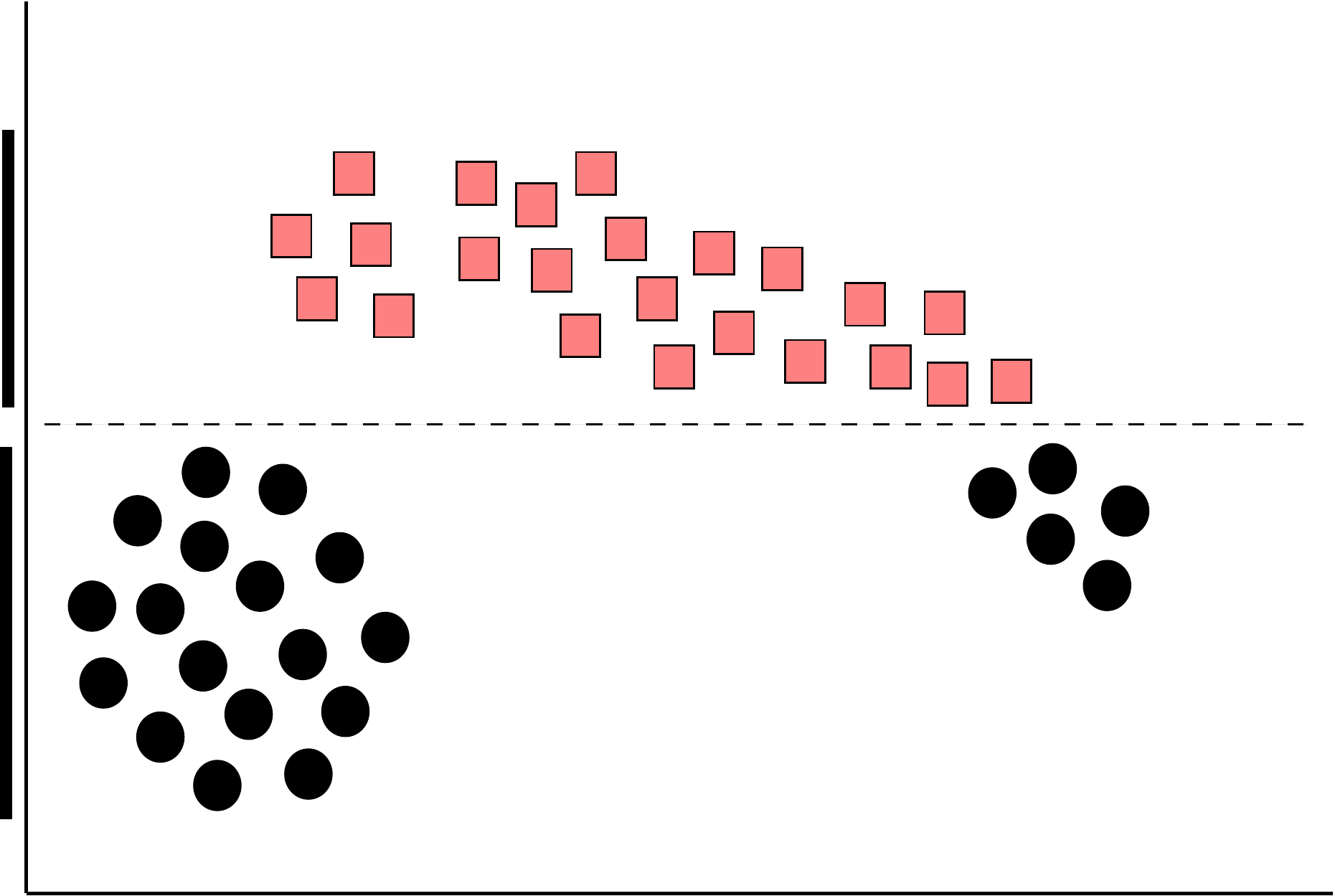}}
\end{center}
\vspace{-.05in}
\caption{\sffamily Why spatially-aware distances are important~\cite{icml10}: first and second partitions are equidistant from the reference partition under a set-based distance. However, \textit{FP} is clearly more similar to \textit{RP} than \textit{SP} (See \textit{2D2C} in Table~\ref{tbl:nonspatial}).\label{fig:bad}}
\end{figure}

These methods ignore the actual spatial description of the data, merely treating the data as atoms in a set and using set information to compare the partitions. As has been observed by many researchers~\cite{zhou,adco,icml10}, ignoring the spatial relationships in the data can be problematic.  Consider the three partitions in Figure~\ref{fig:bad}. The first partition (FP) is obtained by a projection onto the $y$-axis, and the second (SP) is obtained via a projection onto the $x$-axis. Partitions (FP) and (SP) are both equidistant from partition (RP) under any of the above mentioned distances, and yet it is clear that (FP) is more similar to the reference partition, based on the spatial distribution of the data.

Some researchers have proposed spatially-aware distances between partitions~\cite{zhou,adco,icml10} (as we review below in Section~\ref{sssec:prior-spat-aware}), but they all suffer from various deficiencies. They compromise the spatial information captured by the clusters (\cite{zhou,adco}), 
they lack metric properties (\cite{zhou,icml10}) (or have discontinuous ranges of distances to obtain metric properties (\cite{adco})), or
they are expensive to compute, making them ineffective for large data sets (\cite{icml10}).  

\vspace{-.1in}

\subsection{Our Work.}
In this paper we exploit a concise, linear reproducing kernel Hilbert space (RKHS) representation of clusters. 
We use this representation to construct an efficient spatially-aware metric between partitions and an efficient spatially-aware consensus clustering algorithm.  	

We build on some recent ideas about clusters: 
(i) that a cluster can be viewed as a sample of data points from a distribution~\cite{JGSSvL09}, 
(ii) that through a similarity kernel $K$, a distribution can be losslessly lifted to a single vector in a RKHS~\cite{muller} and 
(iii) that the resulting distance between the representative vectors of two distributions in the RKHS can be used as a metric on the distributions~\cite{muller,hilbert,smola2007hilbert}.  

\paragraph{Representations.} We first adapt the representation of the clusters in the RKHS in two ways: approximation and normalization.  
Typically, vectors in an RKHS are infinite dimensional, but they can be approximated arbitrarily well in a finite-dimensional $\ell_2$ space that retains the linear structure of the RKHS~\cite{Rahimi2007,current}.  This provides concise and easily manipulated representations for entire clusters.  
Additionally, we normalize these vectors to focus on the spatial information of the clusters.  This turns out to be important in consensus clustering, as illustrated in Figure \ref{fig:normal}. 

\paragraph{Distance Computation.} Using this convenient representation (an approximate normalized RKHS vector), we develop a metric between partitions.  Since the clusters can now be viewed as points in (scaled) Euclidean space we can apply standard measures for comparing point sets in such spaces.  In particular, we define a spatially-aware metric \liftemd between partitions as the transportation distance~\cite{emd} between the representatives, weighted by the number of points they represent.  While the  transportation distance is a standard distance metric on probability distributions, it is expensive to compute (requiring $O(n^3)$ time for $n$ points)~\cite{hungarian}. However, since the points here are clusters, and the number of clusters ($k$) is typically significantly less than the data size ($n$), this is not a significant bottleneck as we see in Section~\ref{sec:expt}.

\paragraph{Consensus.} We exploit the linearity of the RKHS representations of the clusters to design  an efficient consensus clustering algorithm.  Given several partitions, each represented as a set of vectors in an RKHS, we can find a partition of this data using standard Euclidean clustering algorithms.  In particular, we can compute a consensus partition by simply running $k$-means (or hierarchical agglomerative clustering) on the lifted representations of each cluster from all input partitions. This reduction from consensus to Euclidean clustering is a key contribution: it allows us to utilize the extensive research and fast algorithms for Euclidean clustering, rather than designing complex hypergraph partitioning methods~\cite{nmi}.

\paragraph{Evaluation.} All of these aspects of our technical contributions are carefully evaluated on real-world and synthetic data.  As a result of the convenient isometric representation, the well-founded metric, and reduction to many existing techniques, our methods perform well compared to previous approaches and are much more efficient.  

\vspace{-.05in}

\subsection{Background}
\label{sec:back}

\subsubsection{Clusters as Distributions.}
The core idea in doing spatially aware comparison of partitions is to treat a cluster as a distribution over the data, for example as a sum of $\delta$-functions at each point of the cluster~\cite{icml10} or as a spatial density over the data~\cite{adco}. The distance between two clusters can then be defined as a distance between two distributions over a metric space (the underlying spatial domain). 
\vspace{-.05in}

\subsubsection{Metrizing Distributions.}

There are standard constructions for defining such a distance; the most well known metrics are the transportation distance~\cite{emd} (also known as the Wasserstein distance, the Kantorovich distance, the Mallows distance or the Earth mover's distance), and the Prokhorov metric~\cite{prokhorov}. Another interesting approach was initiated by M\"{u}ller~\cite{muller}, and develops a metric between general measures based on integrating test functions over the measure. When the test functions are chosen from a reproducing kernel Hilbert space~\cite{Aronszajn1950} (RKHS), the resulting metric on distributions has many nice properties~\cite{hilbert,gretton2008kernel,smola2007hilbert,berlinet2004reproducing}, most importantly that it can be isometrically embedded into the Hilbert space, yielding a convenient (but infinite dimensional) representation of a measure. 

This measure has been applied to the problem of computing a \emph{single} clustering by Jegelka \emph{et. al.}~\cite{JGSSvL09}. In their work, each cluster is treated as a distribution and the partition is found by maximizing the inter-cluster distance of the cluster representatives in the RKHS.  We modify this distance and its construction in our work.

A parallel line of development generalized this idea independently to measures over higher dimensional objects (lines, surfaces and so on). The resulting metric (the \emph{current distance}) is exactly the above metric when applied to $0$-dimensional objects (scalar measures) and has been used extensively~\cite{Vaillant2005,GlaunesJoshi:MFCA:06,DBLP:conf/miccai/DurrlemanPTA08,current} to compare shapes. In fact, thinking of a cluster of points as a ``shape'' was a motivating factor in this work.  
\vspace{-.1in}
\subsubsection{Distances Between Partitions.}
\label{sssec:prior-spat-aware}
Section~\ref{sec:intro} reviews spatially-aware and space-insensitive approaches to comparing partitions. We now describe prior work on spatially-aware distances between partitions in more detail.

\begin{figure}[ht]
\begin{center}
 \includegraphics[width=.6\linewidth]{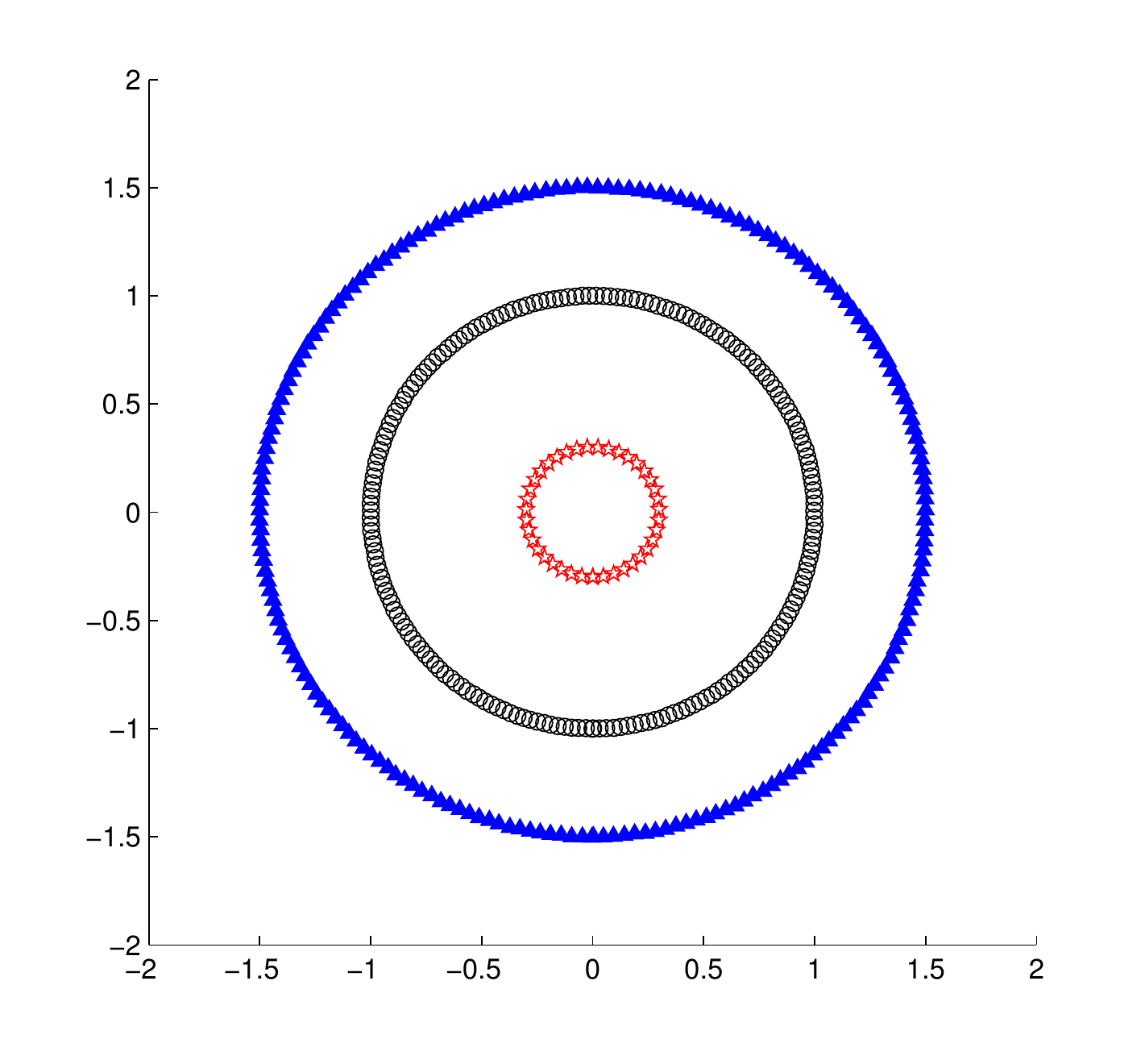}
\end{center}
 \vspace{-.2in}
 \caption{\sffamily Dataset with $3$ concentric circles, each representing a cluster partitioning the data.  The \mallows distance~\cite{zhou} can not distinguish between these clusters.\label{fig:mallowsbad}}
\vspace{-.11in}
\end{figure} 

Zhou~\etal\cite{zhou} define a distance metric \mallows by replacing each cluster by its centroid (this of course assumes the data does not lie in an abstract metric space), and computing a weighted transportation distance between the sets of cluster centroids. 
Technically, their method yields a pseudo-metric, since two different clusters can have the same centroid, for example in the case of concentric ring clusters (Figure~\ref{fig:mallowsbad}).
It is also oblivious to the distribution of points within a cluster. 

Coen~\etal\cite{icml10} avoid the problem of selecting a cluster center by defining the distance between clusters as the transportation distance between the full sets of points comprising each cluster. This yields a metric on the set of all clusters in both partitions. In a second stage, they define the \emph{similarity distance} \cdistance between two partitions as the ratio between the transportation distance between the two partitions (using the metric just constructed as the base metric) and a ``non-informative'' transportation distance in which each cluster center distributes its mass equally to all cluster centers in the other partition. While this measure is symmetric, it does not satisfy triangle inequality and is therefore not a metric. 

Bae~\etal\cite{adco} take a slightly different approach. They build a spatial histogram over the points in each cluster, and use the counts as a vector signature for the cluster. Cluster similarity is then computed via a dot product, and the similarity between two partitions is then defined as the sum of cluster similarities in an optimal matching between the clusters of the two partitions, normalized by the self-similarity of the two partitions.

\begin{figure}[ht]
\begin{center}
  \includegraphics[width=.8\linewidth]{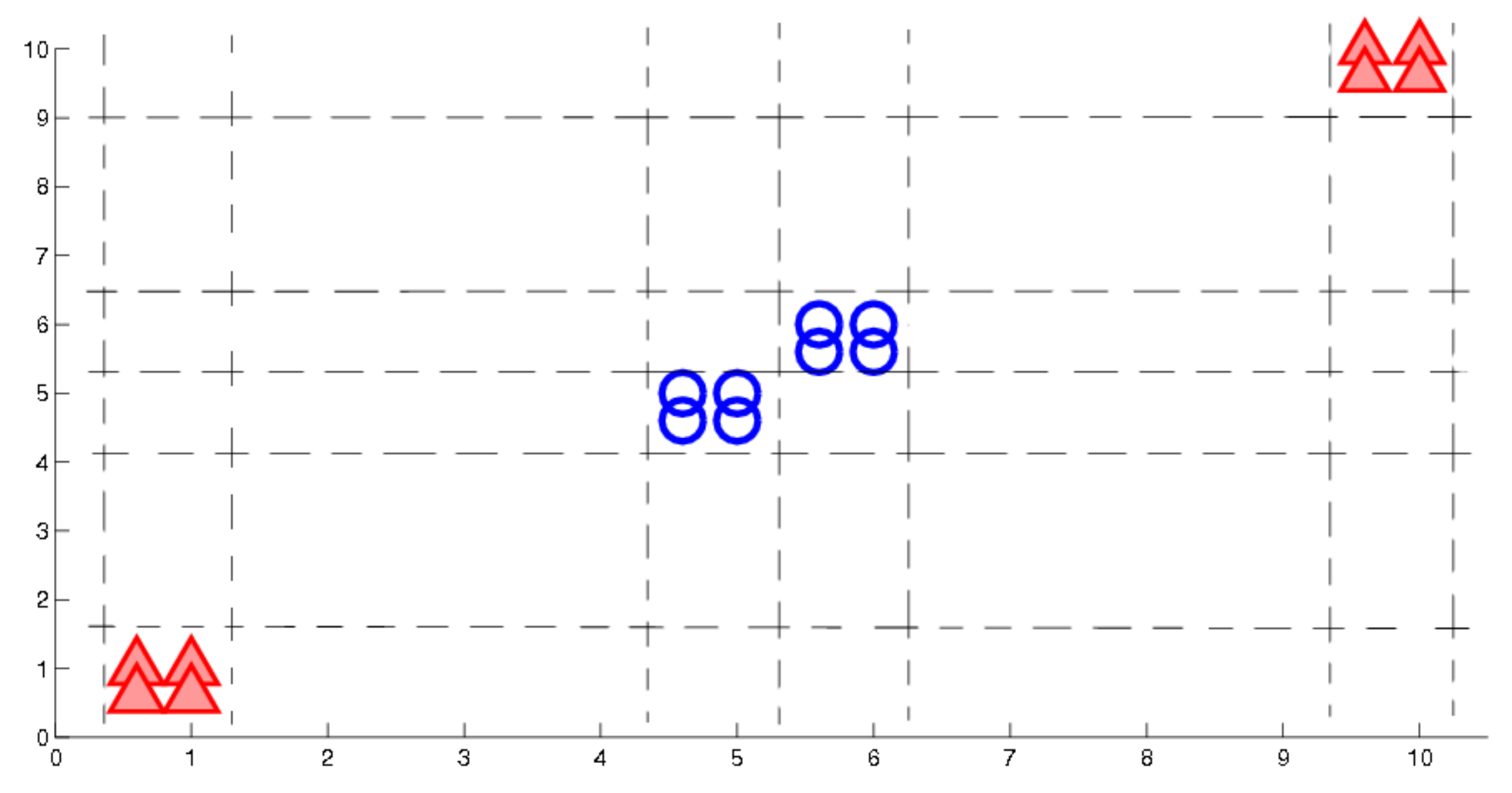}
\end{center}
\vspace{-.2in}
\caption{\sffamily Data set with $4$ clusters, each a set grouped in a single grid cell.  The two clusters with blue open circles are as close as the two clusters with filled red triangles under the \adco distance~\cite{adco}.\label{fig:adcobad}}
\vspace{-.11in}
\end{figure} 

In general, such a spatial partitioning would require a number of histogram bins exponential in the dimension; they get around this problem by only retaining information about the marginal distributions along each dimension. One weakness of this approach is that only points that fall into the same bin contribute to the overall similarity. This can lead dissimilar clusters to be viewed as similar; in Figure~\ref{fig:adcobad}, 
the two $\bigtriangleup$ (red) clusters will be considered as similar as the two $\bigcirc$ (blue) clusters.

Their approach yields a similarity, and not a distance metric. In order to construct a metric, they have to do the usual transformation $\text{dist} = 1 - \text{sim}$ and then add one to each distance between non-identical items, which yields the somewhat unnatural (and discontinuous) metric \adco.  Their method also implicitly assumes (like Zhou \etal\cite{zhou}) that the data lies in Euclidean space.

\paragraph{Our approach.}
Our method, centered around the RKHS-based metric between distributions, addresses all of the above problems. It yields a true metric, incorporates the actual distribution of the data correctly, and avoids exponential dependency on the dimension. The price we pay is the requirement that the data lie in a space admitting a positive definite kernel. However, this actually enables us to apply our method to clustering objects like graphs and strings, for which similarity kernels exist~\cite{graphkernel,stringkernel} but no convenient vector space representation is known. 

\vspace{-.1in}
\subsubsection{Consensus Clustering Algorithms.}
One of the most popular methods for computing consensus between a collection of partitions is the \emph{majority rule}: for each pair of points, each partition ``votes'' on whether the pair of points is in the same cluster or not, and the majority vote wins. While this method is simple, it is expensive and is spatially-oblivious; two points might lie in separate clusters that are close to each other. 

Alternatively, consensus can be defined via a $1$-median formulation: given a distance between partitions, the consensus partition is the partition that minimizes the sum of distances to all partitions. If the distance function is a metric, then the best partition from among the input partitions is guaranteed to be within twice the cost of the optimal solution (via triangle inequality). In general, it is challenging to find an arbitrary partition that minimizes this function. For example, the above majority-based method can be viewed as a heuristic for computing the $1$-median under the Rand distance, and algorithms with formal approximations exist for this problem~\cite{gionis}.

Recently Ansari \etal~\cite{AFC10} extended these above schemes to be spatially-aware by inserting \cdistance in place of Rand distance above.  This method is successful in grouping similar clusters from an ensemble of partitions, but it is quite slow on large data sets $P$ since it requires computing \emd (defined in Section \ref{sec:defn}) on the full dataset.  
Alternatively using representations of each cluster in the ambient space (such as its mean, as in \mallows \cite{zhou}) would produce another spatially-aware ensemble clustering variant, but would be less effective because its representation causes unwanted simplification of the clusters; see Figure \ref{fig:mallowsbad}.  

\section{Preliminaries}

\subsection{Definitions.}
\label{sec:defn}
Let $P$ be the set of points being clustered, with $|P| = n$. We use the term \emph{cluster} to refer to a subset $C$ of $P$ (i.e an actual cluster of the data), and the term \emph{partition} to refer to a partitioning of $P$ into clusters (i.e what one would usually refer to as a clustering of $P$). Clusters will always be denoted by the capital letters $A, B, C, \ldots$, and partitions will be denoted by the symbols $\c{A}, \c{B}, \c{C}, \ldots$. We will also consider \emph{soft partitions} of $P$, which are fractional assignments $\{p(C|x)\}$ of points $x$ to clusters $C$ such that for any $x$, the assignment weights $p(C|x)$ sum to one.

We will assume that $P$ is drawn from a space $X$ endowed with a \emph{reproducing kernel} $\kappa : X \times X \rightarrow \reals$~\cite{Aronszajn1950}. The kernel $\kappa$ induces a Hilbert space $\mathcal{H}_\kappa$ via the \emph{lifting map} $\Phi : X \rightarrow \mathcal{H}_\kappa$, with the property that $\kappa(x,y) = \langle \Phi(x), \Phi(y) \rangle_\kappa$, $\langle \cdot , \cdot\rangle_\kappa$ being the inner product that defines $\mathcal{H}_\kappa$. 

Let $p,q$ be probability distributions defined over $X$. Let $\c{F}$ be a set of real-valued bounded measurable functions defined over $X$. Let $\c{F}_\kappa \triangleq \lbrace f \in \c{F} \mid \|f\|_\kappa \le 1\rbrace$ denote the \emph{unit ball} in the Hilbert space $\mathcal{H}_\kappa$. The \emph{integral probability metric}~\cite{muller} $\gamma_\kappa$ on distributions $p,q$ is defined as 
$
\gamma_\kappa(p,q) = \sup_{f \in \c{F}_\kappa} |\int_X f dp - \int_X f dq |.
$
We will make extensive use of the following explicit formula for $\gamma_\kappa(p,q)$:
\begin{align}
\label{eq:explicit}
\gamma^2_\kappa(p,q) &= \iint_X \kappa(x,y)dp(x)dp(y) 
\\ \nonumber &
+ \iint_X \kappa(x,y)dq(x)dq(y)
\\ \nonumber &
- 2 \iint_X \kappa(x,y)dp(x)dq(y),
\end{align}
which can be derived (via the kernel trick) from the following formula for $\gamma_\kappa$ \cite{hilbert}:
$
\gamma_\kappa(p,q) = \| \int_X \kappa(\cdot, x)dp(x) - \int_X \kappa(\cdot, x)dq(x)\|_{\mathcal{H}_\kappa}. 
$
This formula also gives us the lifting map $\Phi$, since we can write $\Phi(p) = \int_X \kappa(\cdot, x)dp(x)$. 

\paragraph{The transportation metric.}
Let $D : X \times X \rightarrow \reals$ be a metric over $X$. The \emph{transportation distance} between $p$ and $q$ is then defined as 
$
\emd(p,q) = \inf_{f : X \times X \rightarrow [0,1]} \int_X\int_X f(x,y)D(x,y) dx dy,
$
such that $\int_X f(x,y) dx = q(y)$ and $\int_X f(x,y) dy = p(x)$. Intuitively, $f(x,y)D(x,y)$ measures the work in transporting $f(x,y)$ mass from $p(x)$ to $q(y)$. 

\vspace{-.05in}

\subsection{An RKHS Distance Between Clusters.} 
\label{sec:hilbert}

\label{sec:hilbert-example} 
We use $\gamma_\kappa$ to construct a metric on clusters. Let $C \subset P$ be a cluster. We associate with $C$ the distribution $p(C) = \sum_x p(C|x) w(x)\delta_x(\cdot)$, where $\delta_x(\cdot)$ is the Kronecker $\delta$-function and $w : P \rightarrow [0,1]$ is a weight function. Given two clusters $C, C' \subset P$, we define 
$
d(C, C') = \gamma_\kappa(p(C), p(C')). 
$

\paragraph{An example.} 
\emph{A simple example illustrates how this distance generalizes pure partition-based distances between clusters. 
Suppose we fix the kernel $\kappa(x,y)$ to be the discrete kernel: $\kappa(x,x) = 1,\, \kappa(x,y) = 0 \; \forall x \ne y$. 
Then it is easy to verify that $d(C, C') = \sqrt{|C \Delta C'|}$ is the square root of the cardinality of the symmetric difference $C \Delta C'$, which is a well known set-theoretic measure of dissimilarity between clusters. Since this kernel treats all distinct points as equidistant from each other, the only information remaining is the set-theoretic difference between the clusters. As $\kappa$ acquires more spatial information, 
$d(C, C')$ incorporates this into the distance calculation.}
\vspace{-.05in}

\subsubsection{Representations in $\mathcal{H}_\kappa$.}

There is an elegant way to represent points, clusters and partitions in the RKHS $\mathcal{H}_\kappa$. Define the lifting map $\Phi(x) = \kappa(\cdot, x)$. This takes a point $x \in P$ to a vector $\Phi(x)$ in $\mathcal{H}_\kappa$. A cluster $C \subset P$ can now be expressed as a weighted sum of these vectors: $\Phi(C) = \sum_{x \in C} w(x)\Phi(x)$. Note that for clarity, in what follows we will assume without loss of generality that all partitions are hard; to construct the corresponding soft partition-based expression, we merely replace terms of the form $\{x \in C\} = \mathbf{1}_{x\in C}$ by the probability $p(C|x)$. 

$\Phi(C)$ is also a vector in $\mathcal{H}_\kappa$, and we can now rewrite $d(C, C')$ as 
$ d(C, C') = \| \Phi(C) - \Phi(C') \|_{\mathcal{H}_\kappa}$

Finally, a partition $\c{P} = \lbrace C_1, C_2, \ldots C_k\rbrace$ of $P$ can be represented by the set of vectors  $\Phi(\c{P}) = \lbrace \Phi(C_i) \rbrace$ in $\mathcal{H}_\kappa$. We note that as long as the kernel is chosen correctly \cite{hilbert}, this mapping is isometric, which implies that the representation $\Phi(C)$ is a \emph{lossless} representation of $C$. 

The linearity of representation is a crucial feature of how clusters are represented. While in the original space, a cluster might describe an unusually shaped collection of points, the same cluster in $\mathcal{H}_\kappa$ is merely the weighted sum of the corresponding vectors $\Phi(x)$. As a consequence, it is easy to represent \emph{soft partitions} as well. A cluster $C$ can be represented by the vector $\Phi(C) = \sum_x w(x) p(C|x) \Phi(x)$.  

\vspace{-.05in}

\subsubsection{An RHKS-based Clustering.}
\label{sec:RKHS-cluster}

Jegelka \etal~\cite{JGSSvL09} used the RKHS-based representation of clusters to formulate a new cost function for computing a \emph{single} clustering.  In particular, they considered the optimization problem of finding the partition $\c{P} = \{C_1, C_2\}$ of two clusters to maximize
\begin{align*}
C(\c{P}) 
= &
|C_1| \cdot |C_2| \cdot \|\Phi(C_1) - \Phi(C_2) \|^2_{\c{H}_K} 
\\ &+ 
\lambda_1 \|\Phi(C_1)\|^2_{\c{H}_\kappa} + \lambda_2 \|\Phi(C_2)\|^2_{\c{H}_\kappa},
\end{align*}
for various choices of kernel $\kappa$ and regularization terms $\lambda_1$ and $\lambda_2$. They mention that this could then be generalized to find an arbitrary $k$-partition by introducing more regularizing terms. Their paper focuses primarily on the generality of this approach and how it connects to other clustering frameworks, and they do not discuss algorithmic issues in any great detail.

\vspace{-.05in}
\section{Approximate Normalized Cluster Representation}

We adapt the RKHS-based representation of clusters $\Phi(C)$ in two ways to make it more amenable to our meta-clustering goals.  
First, we approximate $\Phi(C)$ to a finite dimensional ($\rho$-dimensional) vector.  This provides a finite representation of each cluster in $\b{R}^\rho$ (as opposed to a vector in the infinite dimensional $\c{H}_\kappa$), it retains linearity properties, and it allows for fast computation of distance between two clusters.  
Second, we normalize $\Phi(C)$ to remove any information about the size of the cluster; retaining only spatial information.  This property becomes critical for consensus clustering.  
\vspace{-.05in}

\subsection{Approximate Lifting Map $\tilde \Phi$.}
The lifted representation $\Phi(x)$ is the key to the representation of clusters and partitions, and its computation plays a critical role in the overall complexity of the distance computation. For kernels of interest (like the Gaussian kernel), $\Phi(x)$ cannot be computed explicitly, since the induced RKHS is an infinite-dimensional function space. 

However, we can take advantage of the shift-invariance of commonly occurring kernels\footnote{A kernel $\kappa(x,y)$ defined on a vector space is shift-invariant if it can be written as $\kappa(x,y) = g(x-y)$.}.  For these kernels a random projection technique in Euclidean space defines an approximate lifting map $\tilde{\Phi} : X \times X \rightarrow \reals^\rho$ with the property that for any $x,y \in P$, 
\[
\left| \|\tilde{\Phi}(x) - \tilde{\Phi}(y) \|_2 - \|\Phi(x)-\Phi(y) \|_{\mathcal{H}_\kappa} \right| \le \eps,
\]
where $\eps > 0$ is an arbitrary user defined parameter, and $\rho = \rho(\eps)$. Notice that the approximate lifting map takes points to $\ell_2^\rho$ with the standard inner product, rather than a general Hilbert space.

The specific construction is due to Rahimi and Recht~\cite{Rahimi2007} and analyzed by Joshi \etal~\cite{current}, to yield the following result:

\begin{theorem}[\cite{current}]
\label{thm:currents}
  Given a set of $n$ points $P \subset X$, shift-invariant kernel $\kappa : X \times X \rightarrow \reals$ and any $\eps > 0$, there exists a map $\tilde{\Phi} : X \times X \rightarrow \reals^\rho$, $\rho = O((1/\eps^2) \log n)$, such that for any $x,y \in P$, 
  $\left| \|\tilde{\Phi}(x) - \tilde{\Phi}(y) \|_2 - \|\Phi(x)-\Phi(y) \|_{\mathcal{H}_\kappa} \right| \le \eps$
\end{theorem}

The actual construction is randomized and yields a $\tilde{\Phi}$ as above with probability $1-\delta$, where $\rho = O((1/\eps^2)\log (n/\delta))$. For any $x$, constructing the vector $\tilde{\Phi}(x)$ takes $O(\rho)$ time. 

\subsection{Normalizing $\Phi(C)$.}
\label{lifting}

\begin{figure}[h!!!]
\includegraphics[scale=1.2]{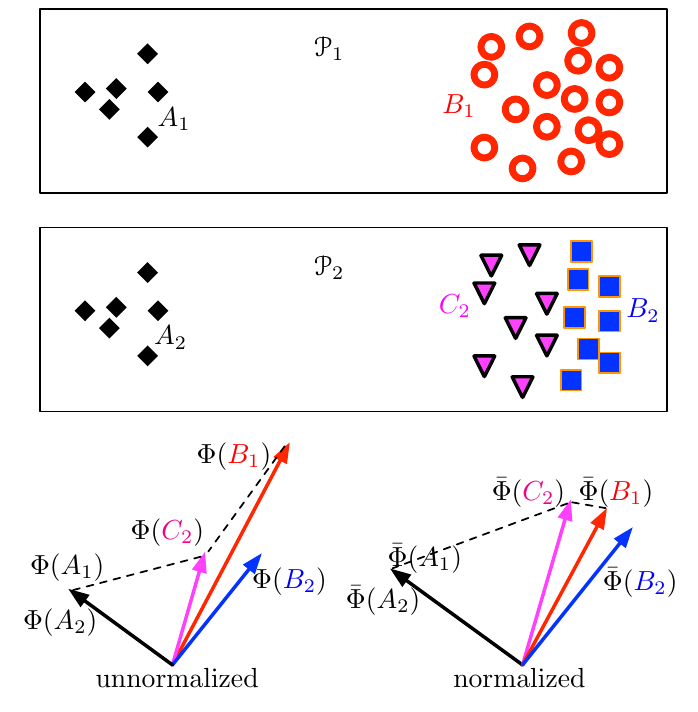}
 \vspace{-.2in}
\caption{\label{fig:normal}\sffamily Two partitions $\c{P}_1 = \{A_1, B_1\}$ and $\c{P}_2 = \{A_2, B_2, C_2\}$ of the same dataset, and a 2-d visualization of all of their representations in an RKHS.  Note that the unnormalized vectors (on the left) have $\Phi(B_1)$ far from $\Phi(B_2)$ and $\Phi(C_2)$ even though the second two are subsets of the first.  The normalized vectors (on the right) have $\bar \Phi(B_1)$ close to both $\bar \Phi(B_2)$ and $\bar \Phi(C_2)$.  In particular, $\Phi(C_2)$ is closer to $\Phi(A_1)$ than $\Phi(B_1)$, but $\bar \Phi(C_2)$ is much closer to $\bar \Phi(B_1)$ than $\bar \Phi(A_1)$.}
\vspace{-.15in}
\end{figure}

The lifting map $\Phi$ is linear with respect to the weights of the data points, while being nonlinear in their location. Since $\Phi(C) = \sum_{x \in C} w(x)\Phi(x)$, this means that any scaling of the vectors $\Phi(x)$ translates directly into a uniform scaling of the weights of the data, and does not affect the spatial positioning of the points. This implies that we are free to normalize the cluster vectors $\Phi(C)$, so as to remove the scale information, retaining only, and exactly, the spatial information. 
In practice, we will normalize the cluster vectors to have unit length; let 
\[
\bar \Phi(C) = \Phi(C) / \|\Phi(C)\|_{\c{H}_\kappa}.
\]
Figure \ref{fig:normal} shows an example of why it is important to compare RKHS representations of vectors using only their spatial information.  In particular, without normalizing, small clusters $C$ will have small norms $\|\Phi(C)\|_{\c{H}_\kappa}$, and the distance between two small vectors $\|\Phi(C_1) - \Phi(C_2)\|_{\c{H}_\kappa}$ is at most $\|\Phi(C_2)\|_{\c{H}_\kappa} + \|\Phi(C_2)\|_{\c{H}_\kappa}$.  Thus all small clusters will likely have similar unnormalized RKHS vectors, irrespective of spatial location.

\vspace{-.1in}

\subsection{Computing the Distance Between Clusters.}
\label{sec:algo}

For two clusters $C, C'$, we defined the distance between them as $d(C, C') = \gamma_\kappa(p(C), p(C'))$. Since the two distributions $p(C)$ and $p(C')$ are discrete (defined over $|C|$ and $|C'|$ elements respectively), we can use \eqref{eq:explicit} to compute $d(C, C')$ in time $O(|C| \cdot |C'|)$. While this may be suitable for small clusters, it rapidly becomes expensive as the cluster sizes increase. 

If we are willing to approximate $d(C, C')$, we can use Theorem~\ref{thm:currents} combined with the implicit definition of $d(C, C')$ as $\|\Phi(C) - \Phi(C')\|_{\mathcal{H}_k}$. Each cluster $C$ is represented as a sum of $|C|$ $\rho$-dimensional vectors, and then the $\ell_2$ distance between the resulting vectors can be computed in $O(\rho)$ time. The following approximation guarantee on $d(C, C')$ then follows from the triangle inequality and an appropriate choice of $\eps$. 

\begin{theorem}
  For any two clusters $C, C'$ and any $\eps > 0$, $d(C, C')$ can be approximated to within an additive error $\eps$ in time $O((|C| + |C'|)\rho)$ time, where $\rho = O((1/\eps^2) \log n)$. 
\end{theorem}

\section{New Distances between Partitions}

Let $\c{P} = \{C_1, C_2, \ldots\}$ and $\c{P}' = \{C'_1, C'_2, \ldots\}$ be two different partitions of $P$ with associated representations $\Phi(\c{P}) = \{\Phi(C_1), \Phi(C_2), \ldots \}$ and $\Phi(\c{P}') = \{\Phi(C'_1), \Phi(C'_2), \ldots\}$. 
Similarly, $\bar \Phi(\c{P}) = \{\bar \Phi(C_1), \bar \Phi(C_2), \ldots \}$ and $\bar \Phi(\c{P}') = \{\bar \Phi(C'_1), \bar \Phi(C'_2), \ldots\}$.
Since the two representations are sets of points in a Hilbert space, we can draw on a number of techniques for comparing point sets from the world of shape matching and pattern analysis. 

We can apply the transportation distance \emd on these vectors to compare the partitions, treating the partitions as distributions. 
In particular, the partition $\c{P}$ is represented by the distribution 
\begin{equation}
\sum_{\bar \Phi(C) \in \bar \Phi(\c{P})} \frac{|C|}{|P|} \cdot \delta_{\bar \Phi(C)}
\label{eq:RKHS-dist}
\end{equation}
where $\delta_{\bar \Phi(C)}$ is a Dirac delta function at $\bar \Phi(C) \in \c{H}_{\kappa}$, with $\mathcal{H}_\kappa$ as the underlying metric. We will refer to this metric on partitions as 
\begin{itemize}
\item $\liftemd(\c{P},\c{P}') = \emd(\bar \Phi(\c{P}),\bar \Phi(\c{P}'))$. 
\end{itemize}


\paragraph{An example, continued.}  
\emph{Once again, we can simulate the loss of spatial information by using the discrete kernel as in Section~\ref{sec:hilbert-example}.  The transportation metric is computed (see Section~\ref{sec:defn}) by minimizing a functional over all partial assignments $f(x,y)$. If we set $f(C, C') = |C \cap C'|/n$ to be the fraction of points overlapping between clusters, then the resulting transportation cost is precisely the Rand distance~\cite{Ran71} between the two partitions! This observation has two implications.  First, that standard distance measures between partitions appear as special cases of this general framework.  Second, $\liftemd(\c{P}, \c{P}')$ will always be at most the Rand distance between $\c{P}$ and $\c{P}'$. }

We can also use other measures. Let 
\[
\vec{\hausdorff}(\Phi(\c{P}), \Phi(\c{P}')) = \max_{v \in \Phi(\c{P})} \min_{w \in \Phi(\c{P}')} \|v - w\|_{\mathcal{H}_\kappa}.
\] 
Then the Hausdorff distance~\cite{hausdorff} is defined as 
\begin{eqnarray*}
\lefteqn{\hausdorff(\Phi(\c{P}), \Phi(\c{P}')) = }
\\ &&
 \max \left(\vec{\hausdorff}\left(\Phi(\c{P}), \Phi(\c{P}')\right), \vec{\hausdorff}\left(\Phi(\c{P}'), \Phi(\c{P})\right)\right). 
\end{eqnarray*}
We refer to this application of the Hausdorff distance to partitions as 
\begin{itemize}
\item $\lifthausdorff(\c{P},\c{P}') = \hausdorff(\bar \Phi(\c{P}),\bar \Phi(\c{P}'))$. 
\end{itemize}

We could also use our lifting map again. 
Since we can view the collection of points $\Phi(\c{P})$ as a spatial distribution in $\mathcal{H}_\kappa$ (see (\ref{eq:RKHS-dist})), we can define $\gamma_{\kappa'}$ in this space as well, with $\kappa'$ again given by any appropriate kernel (for example, $\kappa'(v, w) = \exp(- \|v- w\|^2_{\mathcal{H}_{\kappa'}})$. We refer to this metric as 
\vspace{-.05in}
\begin{itemize} \denselist
\item $\liftcd(\c{P},\c{P}') = \gamma_{\kappa'}(\bar \Phi(\c{P}),\bar \Phi(\c{P}'))$.
\end{itemize}

\subsection{Computing the distance between partitions}

The lifting map $\Phi$ (and its approximation $\tilde{\Phi}$) create efficiency in two ways. First, it is fast to generate a representation of a cluster $C$ ($O(|C|\rho)$ time), and second, it is easy to estimate the distance between two clusters ($O(\rho)$ time). This implies that after a linear amount of processing, all distance computations between partitions depend only on the number of clusters in each partition, rather than the size of the input data. Since the number of clusters is usually orders of magnitude smaller than the size of the input, this allows us to use asymptotically inefficient algorithms on $\tilde \Phi(\c{P})$ and $\tilde \Phi(\c{P}')$ that have small overhead, rather than requiring more expensive (but asymptotically cheaper in $k$) procedures. Assume that we are comparing two partitions $\c{P}, \c{P}'$ with $k$ and $k'$ clusters respectively. \liftemd is computed in general using a min-cost flow formulation of the problem, which is then solved using the Hungarian algorithm. This algorithm takes time $O((k+k')^3)$. While various approximations of \emd exist~\cite{shirdhonkar,indykthaper}, the exact method suffices for our setting for the reasons mentioned above.

It is immediate from the definition of \lifthausdorff that it can be computed in time $O(k \cdot k')$ by a brute force calculation. A similar bound holds for exact computation of \liftcd.  While approximations exist for both of these distances, they incur overhead that makes them inefficient for small $k$. 

\section{Computing Consensus Partitions}
\label{sec:consensus}

As an application of our proposed distance between partitions, we describe how to construct a spatially-aware consensus from a collection of partitions.  
This method  reduces the consensus problem to a standard clustering problem, allowing us to leverage the extensive body of work on standard clustering techniques.  
Furthermore, the representations of clusters as vectors in $\b{R}^\rho$ allows for very concise representation of the data, making our algorithms extremely fast and scalable.  

\subsection{A Reduction from Consensus Finding to Clustering.}
Our approach exploits the linearity of cluster representations in $\mathcal{H}_\kappa$, and works as follows. Let $\c{P}_1, \c{P}_2, \ldots \c{P}_m$ be the input (hard or soft) partitions of $P$. Under the lifting map $\Phi$, each partition can be represented by a set of points $\lbrace \Phi(\c{P}_i)\rbrace $ in $\mathcal{H}_\kappa$. Let $Q = \bigcup_i \Phi(\c{P}_i)$ be the collection of these points. 

\begin{defn}
A (soft) consensus $k$-partition of $\c{P}_1, \c{P}_2, \ldots \c{P}_m$ is a partition $\consensus$ of $\bigcup_i \c{P}_i$ into $k$ clusters $\{C_1^*, \ldots, C_k^*\}$ that minimizes the sum of squared distances from each $\bar \Phi(C_{i,j}) \in \bar \Phi(\c{P}_i)$ to its associated $\bar \Phi(C_l^*) \in \consensus$.  Formally, for a set of $k$ vectors $V = \{v_1, \ldots, v_k\} \subset \c{H}_\kappa$ define 
\[
\liftSSD(\{\c{P}_i\}, V) = 
\sum_{C_{i,j} \in \cup_i \c{P}_i}  \frac{|C_{i,j}|}{n} 
\min_{v \in V} \left\| \bar \Phi(C_{i,j}) - v \right\|_{\c{H}_\kappa}^2
\]
and then define \consensus as the minimum such set
\[
\consensus = 
\underset{V^* = \{v_1^*, \ldots, v_k^*\} \in \c{H}_\kappa}{\operatorname{argmin}} 
\liftSSD(\{\c{P}_i\}_i, V^*).
\] 
\end{defn}

How do we interpret $\consensus$? Observe that each element in $Q$ is the lifted representation of some cluster $C_{i,j}$ in some partition $\c{P}_i$, and therefore corresponds to some subset of $P$. Consider now a single cluster in $\consensus$. Since $\c{H}_\kappa$ is linear and \consensus  { } minimizes distance to some set of cluster representatives, it must be in their linear combination.  Hence it can be associated with a weighted subset of elements of $\Phi(P)$, and is hence a soft partition.  It can be made hard by voting each point $x \in P$ to the representative $C_l^* \in \consensus$ for which it has the largest weight.  

\begin{figure}[h!!!]
\includegraphics[scale=1.2]{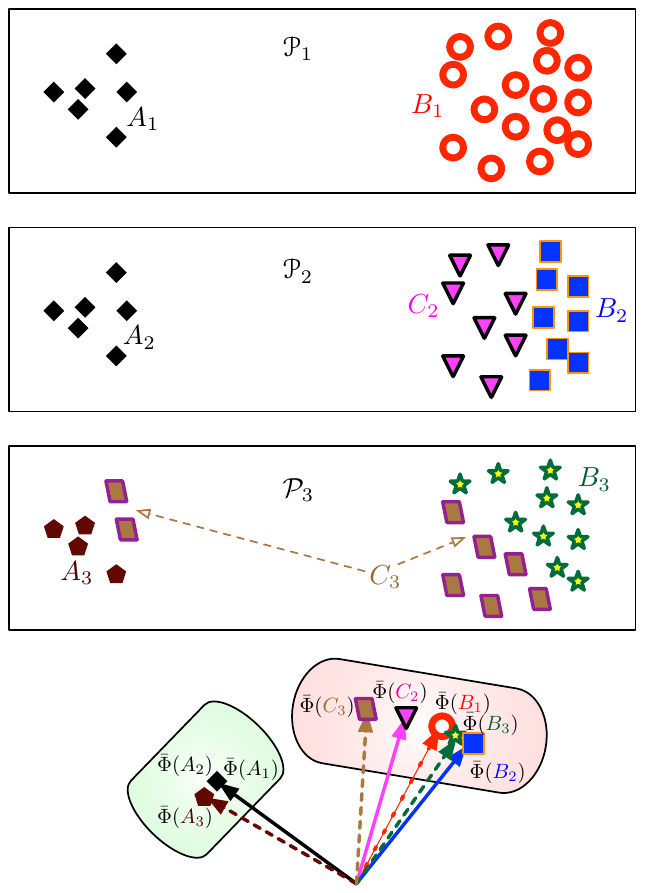}
\label{fig:con-clust}
\caption{\sffamily Three partitions $\c{P}_1 = \{A_1, B_1\}$, $\c{P}_2 = \{A_2, B_2, C_2\}$, $\c{P}_3 = \{A_3, B_3, C_3\}$ of the same dataset, and a 2-d visualization of all of their representations in an RKHS.  
These vectors are then clustered into $k=2$ consensus clusters consisting of $\{\bar\Phi(A_1), \bar\Phi(A_2), \bar\Phi(A_3)\}$ and $\{\bar\Phi(B_1), \bar\Phi(B_2), \bar\Phi(C_2), \bar\Phi(B_3), \bar\Phi(C_3)\}$.}
\vspace{-.2in}
\end{figure}

Figure \ref{fig:con-clust} shows an example of three partitions, their normalized RKHS representative vectors, and two clusters of those vectors.  

\subsection{Algorithm.}
\label{kmeans}

We will use the approximate lifting map $\tilde{\Phi}$ in our procedure. This allows us to operate in a $\rho$-dimensional Euclidean space, in which there are many clustering procedures we can use. For our experiments, we will use both $k$-means and hierarchical agglomerative clustering (HAC).  
That is, 
let \liftkmeans be the algorithm of running $k$-means on $\bigcup_i \tilde \Phi (\c{P}_i)$, and 
let \lifthac be the algorithm of running HAC on $\bigcup_i \tilde \Phi(\c{P}_i)$.  For both algorithms, the output is the (soft) clusters represented by the vectors in $\Phi(\consensus)$.  
Our results will show that the particular choice of clustering algorithm (e.g. \liftkmeans or \lifthac) is not crucial.  
There are multiple methods to choose the right number of clusters and we can employ one of them to fix $k$ for our consensus technique. 
Algorithm~\ref{alg:liftCluster} summarizes our consensus procedure.

\begin{algorithm}
\textbf{Input:} (soft) Partitions $\c{P}_1, \c{P}_2, \ldots \c{P}_m$ of $P$, kernel function $\kappa$
\\
\textbf{Output:} Consensus (soft) partition \consensus
  \begin{algorithmic}[1]
    \STATE Set $Q = \cup_i \tilde{\Phi}(\c{P}_i)$
    \STATE Compute $V^* = \lbrace v_1^*, \ldots v_k^* \rbrace \subset \c{H}_\kappa$ to minimizes $\liftSSD(Q,V^*)$.
    \\ (Via $k$-means for \liftkmeans or HAC for \lifthac) 
    \STATE Assign each $p \in P$ to the cluster $C_i \in \consensus$ associated with the vector $v^*_i \in V^*$ 
    \begin{itemize}\denselist \vspace{-.08in}
    \item for which $\langle \tilde\Phi(p), v_i \rangle$ is maximized for a hard partition, or
     \item with weight proportional to $\langle \tilde \Phi(p), v_i \rangle$ for a soft partition.
     \end{itemize}
    \STATE Output $\consensus$.
    \end{algorithmic}
  \caption{\sffamily Consensus finding\label{alg:liftCluster}}
\end{algorithm}

\paragraph{Cost analysis.} 

Computing $Q$ takes $O(mn\rho) = O(mn\log n)$ time. Let $|Q|=s$. Computing $\consensus$ is a single call to any standard Euclidean algorithm like $k$-means that takes time $O(sk\rho)$ per iteration, and computing the final soft partition takes time linear in $n(\rho + k) + s$. Note that in general we expect that $k, s \ll n$.
In particular, when $s < n$ and $m$ is assumed constant, then the runtime is $O(n (k + \log n))$.  

\vspace{-.1in}
\section{Experimental Evaluation}
\label{sec:expt}

In this section we empirically show the effectiveness of our distance between partitions, \liftemd and \liftcd, and the consensus clustering algorithms that conceptually follow, \liftkmeans and \lifthac.

\paragraph*{Data.} 
We created two synthetic datasets in $\reals^2$ namely, \textit{2D2C} for which data is drawn from $2$ Gaussians to produce $2$ visibly separate clusters and \textit{2D3C} for which the points are arbitrarily chosen to produce $3$ visibly separate clusters. We also use 5 different datasets from the UCI repository~\cite{uci} (Wine, Ionosphere, Glass, Iris, Soybean) with various numbers of dimensions and labeled data classes. To show the ability of our consensus procedure and the distance metric to handle large data, we use both the training and test data of MNIST~\cite{mnist} database of handwritten digits which has 60,000 and 10,000 examples respectively in  $\reals^{784}$. 

\paragraph*{Methodology.} We will compare our approach with the partition-based measures namely, Rand Distance and Jaccard distance and information-theoretic measures namely, Normalized Mutual Information and Normalized Variation of Information \cite{Wagner07comparingclusterings}, as well as the spatially-aware measures \adco\cite{adco} and \cdistance\cite{icml10}. We ran $k$-means, single-linkage, average-linkage, complete-linkage and Ward's method~\cite{datamining} on the datasets to generate input partitions to the consensus clustering methods. We use accuracy~\cite{DBLP:conf/icml/DingL07} and Rand distance~\cite{Ran71} to measure the effectiveness of the consensus clustering algorithms by comparing the returned consensus partitions to the original class labeling. We compare our consensus technique against few hypergraph partitioning based consensus methods CSPA, HGPA and MCLA~\cite{nmi}. For the MNIST data we also visualize the cluster centroids of each of the input and consensus partitions in a 28x28 grayscale image.

Accuracy studies the one-to-one relationship between clusters and classes; it measures the extent to which each cluster contains data points from the corresponding class.  Given a set $P$ of $n$ elements, consider a set of $k$ clusters $\c{P} = \{C_1, \ldots, C_k\}$ and $m\geq k$ classes $\c{L} = \{L_1, \ldots, L_m\}$ denoting the ground truth partition.  
Accuracy is expressed as 
\[
\textsf{A}(\c{P},\c{L}) = \max_{\mu : [1:k] \to [1:m]} \sum_{i=1}^k \frac{|C_i \cap L_{\mu(i)}|}{n},
\] 
where $\mu$ assigns each cluster to a distinct class.  
The Rand distance counts the fraction of pairs which are assigned consistently in both partitions as in the same or in different classes.  
Let $R_S(\c{P}, \c{L})$ be the number of pairs of points that are in the same cluster in both $\c{P}$ and $\c{L}$, and 
let $R_D(\c{P}, \c{L})$ be the number of pairs of points that are in different clusters in both $\c{P}$ and $\c{L}$.  
Now we can define the Rand distance as
\[
\textsf{R}(\c{P},\c{L}) = 1 - \frac{R_S(\c{P},\c{L}) + R_D(\c{P},\c{L})}{{n \choose 2}}.
\]

\paragraph*{Code.} 
We implement $\tilde \Phi$ as the \emph{Random Projection} feature map~\cite{current} in C to lift each data point into $\reals^\rho$.
We set $\rho = 200$ for the two synthetic datasets \textit{2D2C} and \textit{2D3C} and all the UCI datasets. 
We set $\rho = 4000$ for the larger datasets, MNIST training and MNIST test.
The same lifting is applied to all data points, and thus all clusters. 

The \liftemd, \liftcd, and \lifthausdorff distances between two partitions $\c{P}$ and $\c{P}'$ are computed by invoking brute-force transportation distance, kernel distance, and Hausdorff distance on $\tilde \Phi(\c{P}), \tilde \Phi(\c{P}') \subset \reals^\rho$ representing the lifted clusters.

To compute the consensus clustering in the lifted space, we apply $k$-means (for \liftkmeans) or HAC (for \lifthac) (with the appropriate numbers of clusters) on the set $Q \subset \reals^\rho$ of all lifted clusters from all partitions.  
The only parameters required by the procedure is the error term $\eps$ (needed in our choice of $\rho$) associated with $\tilde \Phi $, and any clustering-related parameters.

We used the cluster analysis functions in MATLAB with the default settings to generate the input partitions to the consensus methods and the given number of classes as the number of clusters. We implemented the algorithm provided by the authors~\cite{adco} in MATLAB to compute \adco. To compute \cdistance, we used the code provided by the authors~\cite{icml10}. We used the ClusterPack MATLAB toolbox ~\cite{clusterpack} to run the hypergraph partitioning based consensus methods CSPA, HGPA and MCLA. 

\subsection{Spatial Sensitivity.} 
\begin{figure}[h]
\begin{center}
\subfigure[Reference Partition (\textit{RP}) \label{fig:2d3c-a}]{\includegraphics[angle=90,width=.3\linewidth]{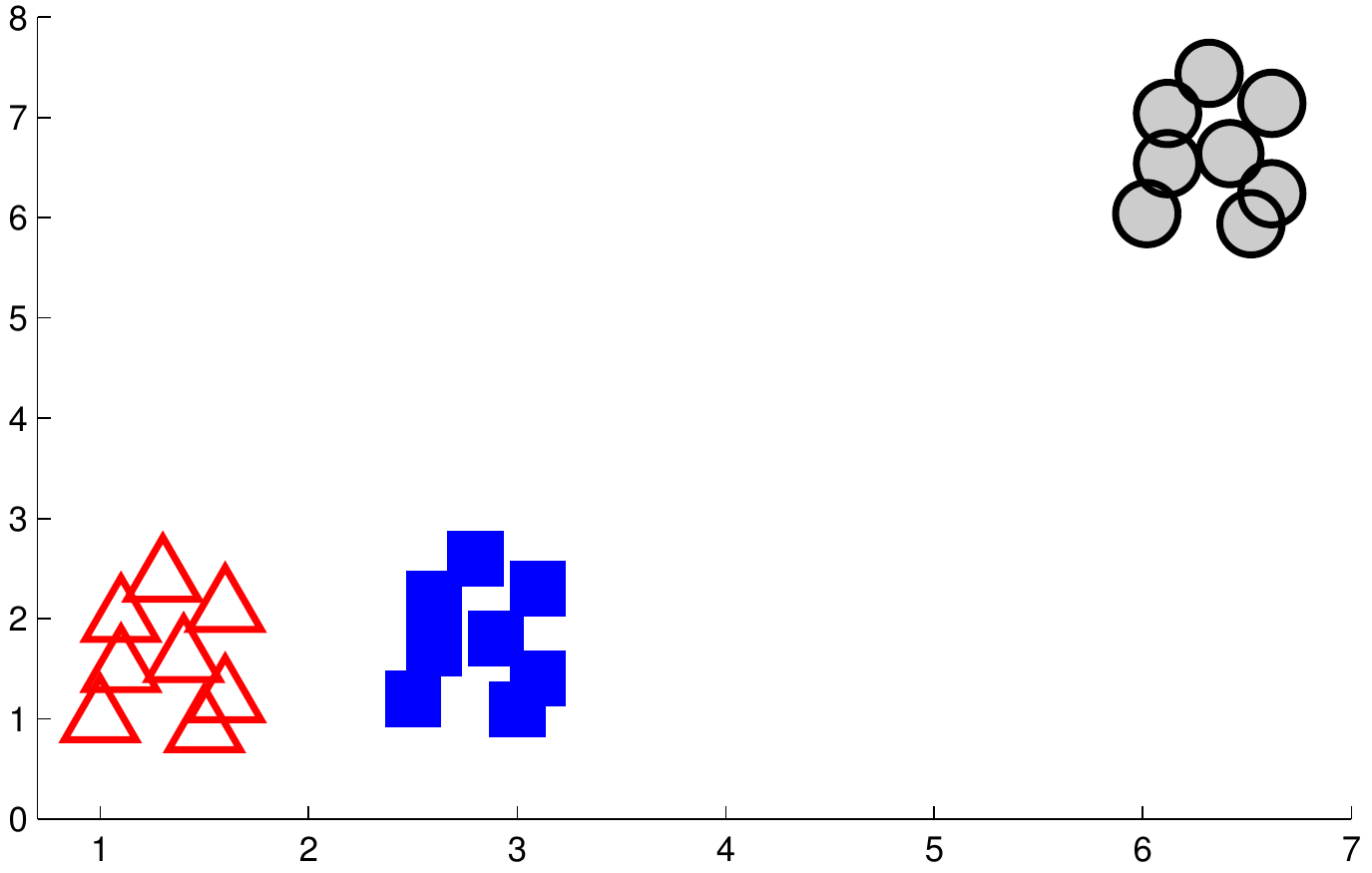}}
\subfigure[First Partition (\textit{FP})\label{fig:2d3c-b}]{\includegraphics[angle=90,width=.3\linewidth]{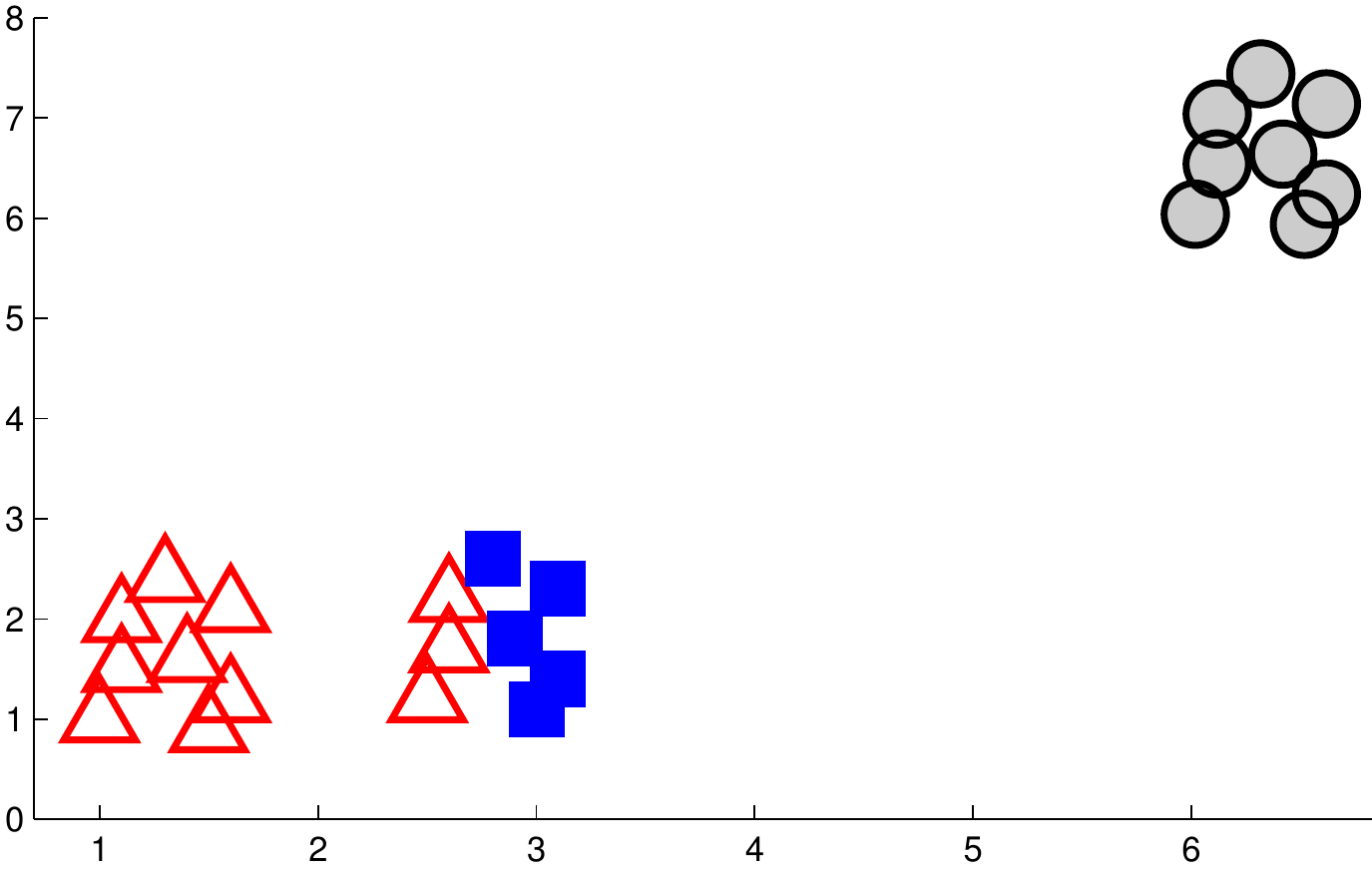}}
\subfigure[Second Partition (\textit{SP})\label{fig:2d3c-c}]{\includegraphics[angle=90,width=.3\linewidth]{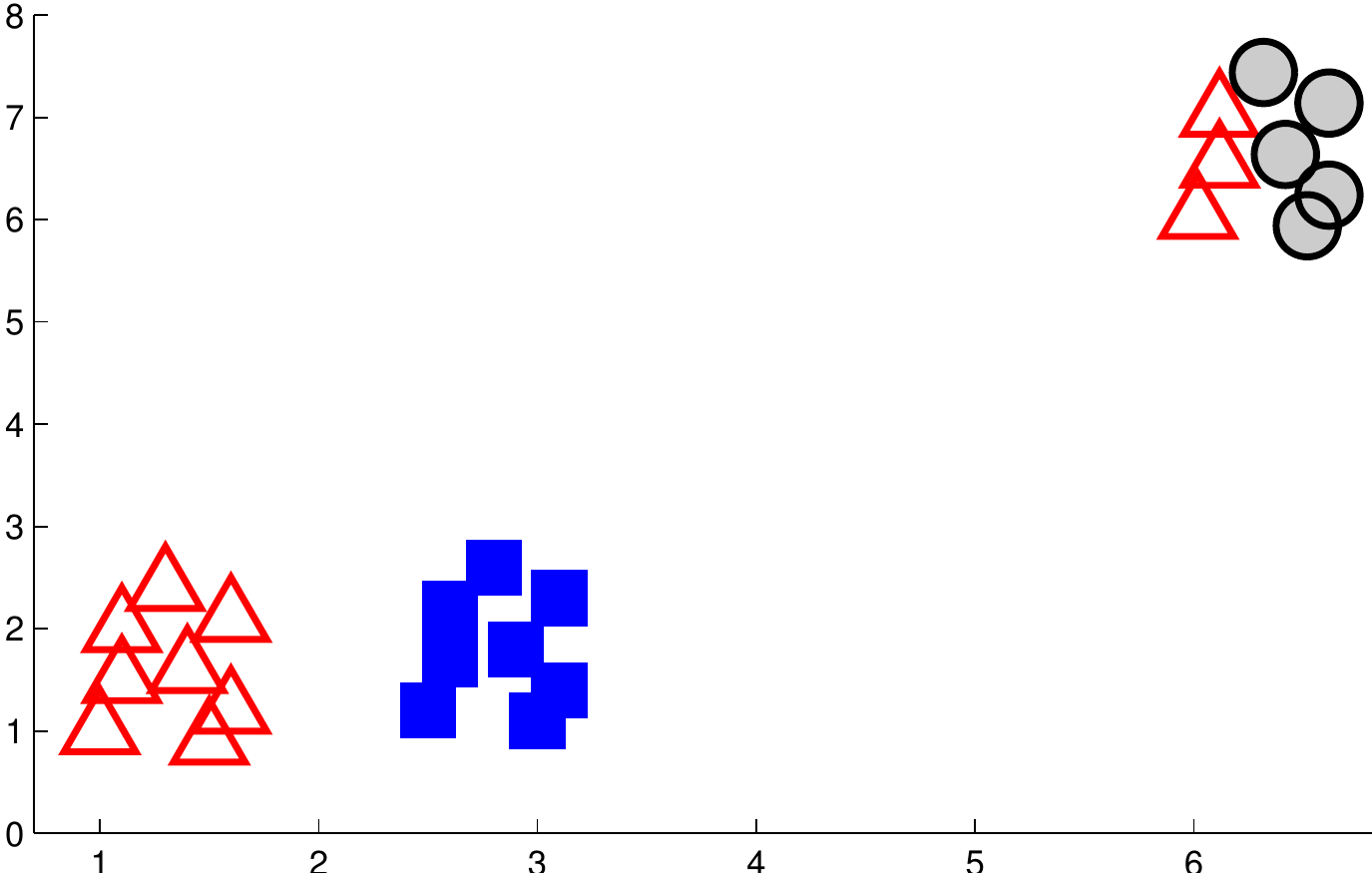}}
\end{center}
\vspace{-.2in}
\caption{\sffamily Different partitions of \textit{2D3C} dataset.\label{fig:2D3C}}
\vspace{-.25in}
\end{figure}

We start by evaluating the sensitivity of our method. We consider three partitions--the ground truth (or) reference partition (\textit{RP}), and manually constructed first and second partitions (\textit{FP} and \textit{SP}) for the datasets \textit{2D2C} (see Figure \ref{fig:bad}) and \textit{2D3C} (see Figure \ref{fig:2D3C}). For both the datasets the reference partition is by construction spatially closer to the first partition than the second partition, but each of the two partitions are equidistant from the reference under any partition-based and information-theoretic measures.  Table \ref{tbl:nonspatial} shows that in each example, our measures correctly conclude that \textit{RP} is closer to \textit{FP} than it is to \textit{SP}. 

\begin{table*}[t]
\small
\begin{center}
  \begin{tabular}{ | l || l | l || l | l || l | l |}
    \hline
    & \multicolumn{2}{|c||}{Dataset \textit{2D2C}} & \multicolumn{2}{|c||}{Dataset \textit{2D3C}} \\
    \cline{2-5}
    Technique & d(\textit{RP},\textit{FP}) & d(\textit{RP},\textit{SP}) & d(\textit{RP},\textit{FP}) & d(\textit{RP},\textit{SP})\\ \hline \hline
    \adco & 1.710 & 1.780 & 1.790 & 1.820  \\ \hline
    \cdistance & 0.240 & 0.350 & 0.092 & 0.407  \\ \hline
    \liftemd & 0.430 & 0.512 & 0.256 & 0.310 \\ \hline
    \liftcd & 0.290 & 0.325 & 0.243 & 0.325 \\ \hline
    \lifthausdorff & 0.410 & 0.490 & 1.227 & 1.291 \\ \hline
  \end{tabular}
\caption{\sffamily Comparing Partitions. Each cell indicates the distance returned under the methods along the rows for the dataset in the column.  Spatially, the left column of each data set (\textit{2D2C} or \textit{2D3C}) should be smaller than the right column; this holds for all $5$ spatial measures/algorithms tested.  In all cases, the two partition-based measures and the two information-theoretic measures yield the same values for d(\textit{RP}, \textit{FP}) and d(\textit{RP}, \textit{SP}), but are not shown.\label{tbl:nonspatial}}                                                                                                                                                            
\end{center}
\vspace{-.08in}
\end{table*}

\subsection{Efficiency.}
We compare our distance computation procedure to \cdistance.  We do not compare against \adco and \mallows because they are not well-founded.  
Both \liftemd and \cdistance compute \emd between clusters after an initial step of either lifting to a feature space or computing \emd between all pairs of clusters.  Thus the proper comparison, and runtime bottleneck, is the initial phase of the algorithms; \liftemd takes $O(n \log n)$ time whereas \cdistance takes $O(n^3)$ time. 
Table \ref{tbl:cons-time} summarizes our results.
For instance, on the \textit{2D3C} data set with $n=24$, our initial phase takes $1.02$ milliseconds, and \cdistance's initial phase takes $2.03$ milliseconds. 
On the Wine data set with $n = 178$, our initial phase takes 6.9 milliseconds, while \cdistance's initial phase takes 18.8 milliseconds.  As the dataset size increases, the advantage of \liftemd over \cdistance becomes even larger. On the MNIST training data with $n = 60,000$, our initial phase takes a little less than 30 minutes, while \cdistance's initial phase takes more than 56 hours.  

\begin{table*}[htbp]
\small
\begin{center}
  \begin{tabular}{ | l | l | l | l | l |}
    \hline
    Dataset & Number of points & Number of dimensions & CDistance & \liftemd  \\ \hline \hline
    \textit{2D3C} & 24 & 2 & 2.03 ms & 1.02 ms \\ \hline
    \textit{2D2C} & 45 & 2 & 4.10 ms & 1.95 ms \\ \hline
    Wine & 178 & 13 & 18.80 ms & 6.90 ms \\ \hline
    MNIST test data & 10,000 & 784 & 1360.20 s & 303.90 s \\ \hline
    MNIST training data & 60,000 & 784 & 202681 s & 1774.20 s \\ \hline
  \end{tabular}
\caption{\sffamily Comparison of runtimes: Distance between true partition and partition generated by $k$-means)\label{tbl:cons-time}}
\end{center}
\vspace{-.08in}
\end{table*}

\subsection{Consensus Clustering.} 
We now evaluate our spatially-aware consensus clustering method. We do this first by comparing our consensus partition to the reference solution based on using the Rand distance (i.e a partition-based measure) in  Table~\ref{tbl:spatVpart-Rand}. Note that for all data sets, our consensus clustering methods (\liftkmeans and \lifthac) return answers that are almost always as close as the best answer returned by any of the hypergraph partitioning based consensus methods CSPA, HGPA, or MCLA.  We get very similar results using the accuracy~\cite{DBLP:conf/icml/DingL07} measure in place of Rand. 

In Table~\ref{tbl:spatVpart-EMD}, we then run the same comparisons, but this time using \liftemd (i.e a spatially-aware measure). Here, it is interesting to note that in all cases (with the slight exception of Ionosphere) the distance we get is smaller than the distance reported by the hypergraph partitioning based consensus methods, indicating that our method is returning a consensus partition that is spatially closer to the true answer. 
The two tables also illustrate the flexibility of our framework, since the results using \liftkmeans and \lifthac are mostly identical (with one exception being IRIS under \liftemd). 

To summarize, our method provides results that are comparable or better on partition-based measures of consensus, and are superior using spatially-aware measures. The running time of our approach is comparable to the best hypergraph partitioning based approaches, so using our consensus procedure yields the best overall result. 

\begin{table*}[htbp]
\small
\begin{center}
  \begin{tabular}{ | l |  l  l  l | l  l |}
    \hline
    Dataset &  CSPA & HGPA & MCLA & \liftkmeans & \lifthac \\ \hline \hline
IRIS 	&	\textbf{0.088}	&	0.270	&	0.115	&	\textbf{0.114}	&	0.125	\\ \hline
Glass 	&	\textbf{0.277}	&	\textbf{0.305}	&	0.428	&	0.425	&	0.430	\\ \hline
Ionosphere &	0.422	&	0.502	&	\textbf{0.410}	&	0.420	&	\textbf{0.410}	\\ \hline
Soybean &	0.188	&	\textbf{0.150}	&	0.163	&	\textbf{0.150}	&	0.154	\\ \hline
Wine 	&	\textbf{0.296}	&	0.374	&	0.330	&	0.320	&	\textbf{0.310}	\\ \hline
MNIST test data & 0.149 &	-	& 	0.163	&	\textbf{0.091}   &	\textbf{0.110}	\\ \hline
  \end{tabular}
\caption{\sffamily Comparison of \liftkmeans and \lifthac with hypergraph partitioning based consensus methods under the Rand distance (with respect to ground truth).
The numbers are comparable across each row corresponding to a different dataset, and smaller numbers indicate better accuracy. The top two methods for each dataset are highlighted.
\label{tbl:spatVpart-Rand}}                                                 
\end{center}
\end{table*}

\begin{table*}[htbp]
\small
\begin{center}
  \begin{tabular}{ | l | l  l  l | l  l |}
    \hline
Dataset & 	CSPA 	& HGPA 	& MCLA 	& \liftkmeans 	& \lifthac 	\\ \hline \hline
IRIS 	& 	\textbf{0.113}	& 0.295 & 0.812 & \textbf{0.106} 	& 0.210		\\ \hline
Glass 	& 	0.573 	& \textbf{0.519} & 0.731 & \textbf{0.531}  	& 0.540		\\ \hline
Ionosphere & 	\textbf{0.729} 	& 0.767 & 0.993 & 0.731 	& \textbf{0.720}		\\ \hline
Soybean & 	0.510 	& 0.495 & 0.951 & \textbf{0.277} 	& \textbf{0.290}  	\\ \hline
Wine 	& 	0.873 	& 0.875 & 0.917 & \textbf{0.831} 	& \textbf{0.842} 	\\ \hline
MNIST test data& 0.182 	& - 	& 0.344 & \textbf{0.106} 	& \textbf{0.112} 	\\ \hline
  \end{tabular}
\caption{\sffamily Comparison of \liftkmeans and \lifthac with hypergraph partitioning based consensus methods under \liftemd (with respect to ground truth).
The numbers are comparable across each row corresponding to a different dataset, and smaller numbers indicate better accuracy. The top two methods for each dataset are highlighted.
\label{tbl:spatVpart-EMD}}
\end{center}
\end{table*}

\begin{figure}[h]
\includegraphics[width=\linewidth]{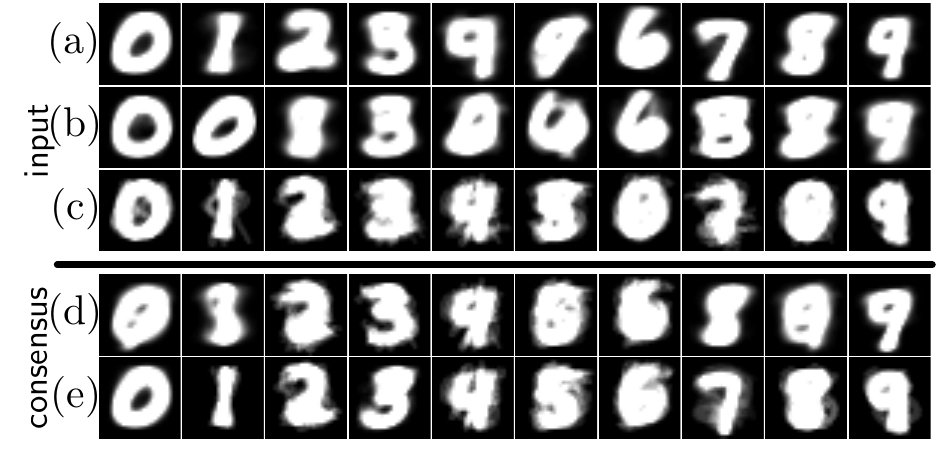}
\vspace{-.2in}
\caption{\sffamily 28x28 pixel representation of the cluster centroids for MNIST test input partitions generated using (a) $k$-means, (b) complete linkage HAC, and (c) average linkage HAC, and the consensus partitions generated by (d) CSPA and (e) \liftkmeans.\label{fig:mnistcentroids}}
\vspace{-.3in}
\end{figure}

We also run consensus experiments on the MNIST test data and compare against CSPA and MCLA. We do not compare against HGPA since it runs very slow for the large MNIST datasets ($n=10,000$); it has quadratic complexity in the input size, and in fact, the authors do not recommend this for large data. 
Figure~\ref{fig:mnistcentroids} provides a visualization of the cluster centroids of input partitions generated using $k$-means, complete linkage HAC and average linkage HAC and the consensus partitions generated by CSPA and \liftkmeans. 
From the $k$-means input, only 5 clusters can be easily associated with digits (\textit{0, 3, 6, 8, 9});
from the complete linkage HAC input, only 7 clusters can be easily associated with digits (\textit{0, 1	, 3, 6, 7, 8, 9}); and 
from the average linkage HAC output, only 6 clusters can be easily associated with digits (\textit{0, 1, 2, 3, 7, 9}). 
The partition that we obtain from running CSPA also lets us identify up to 6 digits (\textit{0, 1, 2, 3, 8, 9}). In all the above three partitions, there occurs cases where two clusters seem to represent the same digit.  In contrast, we can identify 9 digits (\textit{0, 1, 2, 3, 4, 5, 7, 8, 9}) with only the digit \textit{6} being noisy from our \liftkmeans output.

\subsection{Error in $\tilde{\Phi}$.}
\begin{figure}[htbp]
\begin{center}
    \includegraphics[width=.8\linewidth]{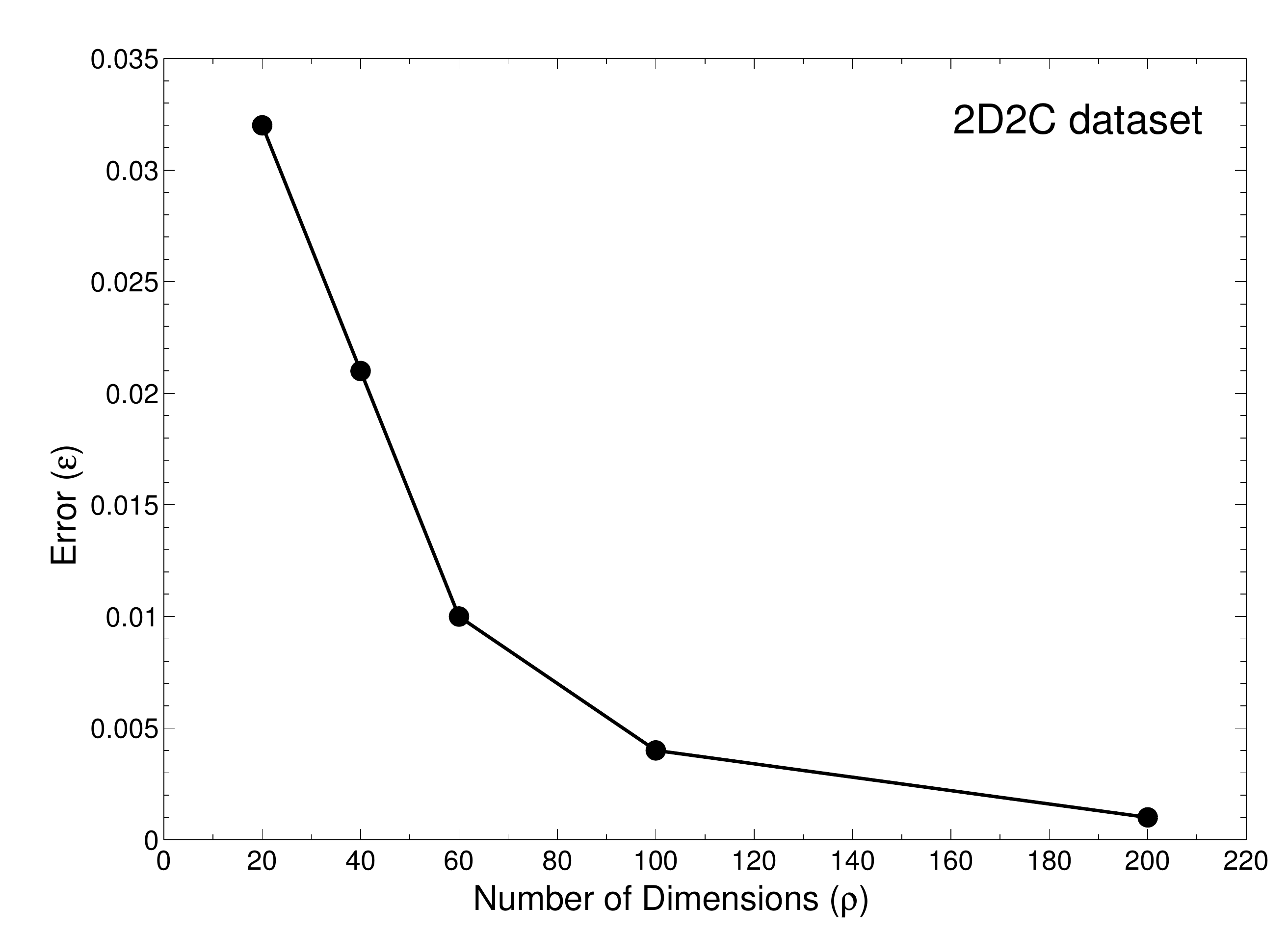}
\end{center}
    \caption{\sffamily Error in \liftemd on \textit{2D2C} dataset (45 samples) as a function of $\rho$. \label{fig:precision2d2c}}
\end{figure}

\begin{figure}[htbp]
\begin{center}
\includegraphics[width=.8\linewidth]{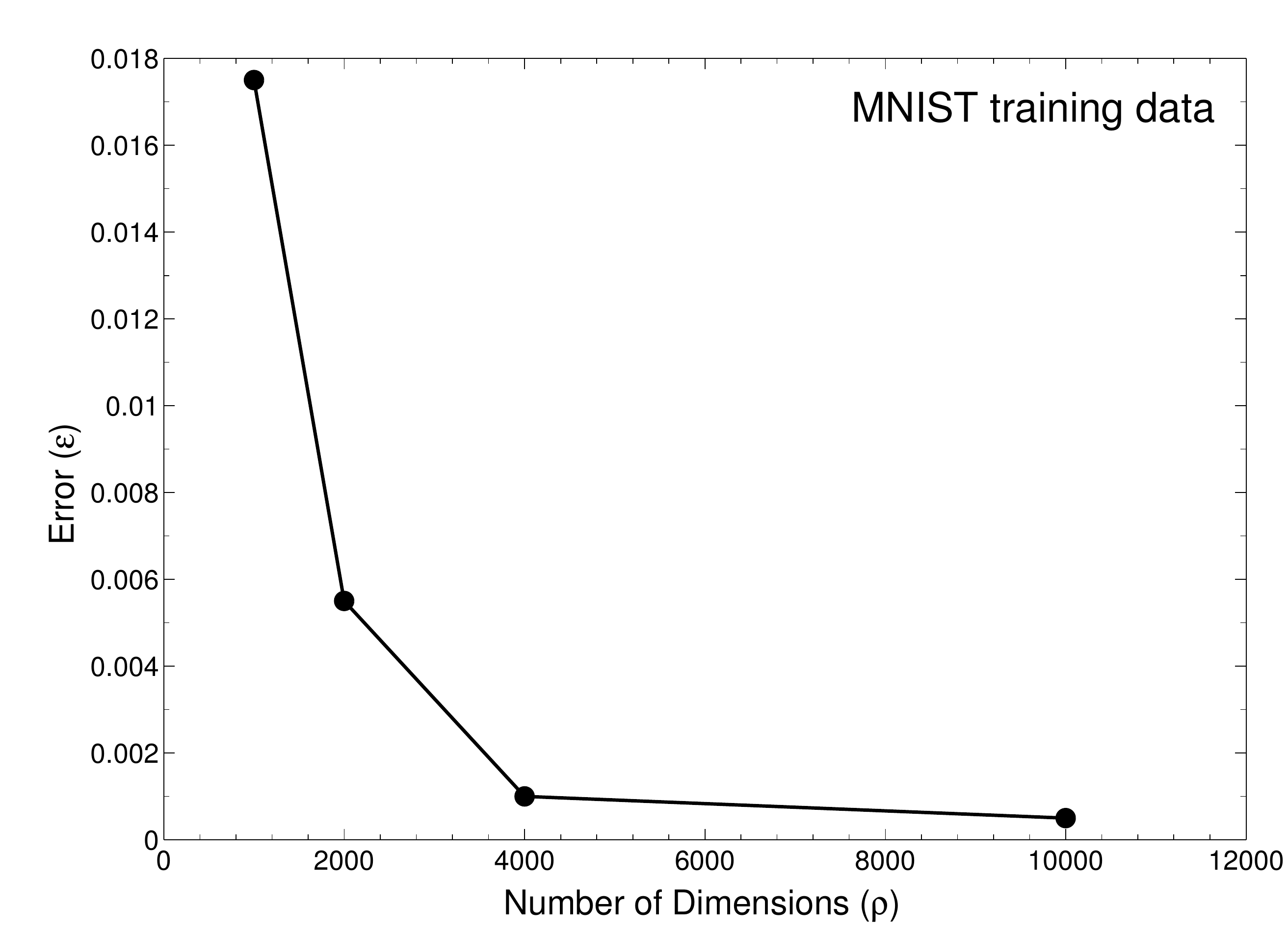}
\end{center}
 \caption{\sffamily Error in \liftemd on MNIST training data (60,000 samples) as a function of $\rho$.  \label{fig:precisionmnist}}
\end{figure}

There is a tradeoff between the desired error $\eps$ in computing \liftemd and the number of dimensions $\rho$ needed for $\tilde{\Phi}$. 
Figure~\ref{fig:precision2d2c} shows the error as a function of $\rho$ on the \textit{2D2C} dataset ($n = 45$).  From the chart, we can see that $\rho = 100$ dimensions suffice to yield a very accurate approximation for the distances. Figure~\ref{fig:precisionmnist} shows the error as a function of $\rho$ on the MNIST training dataset that has $n=60,000$ points.  From the chart, we can see that $\rho = 4,000$ dimensions suffice to yield a very accurate approximation for the distances. 

\section{Conclusions}
We provide a spatially-aware metric between partitions based on a RKHS representation of clusters.  We also provide a spatially-aware consensus clustering formulation using this representation that reduces to Euclidean clustering.  
We demonstrate that our algorithms are efficient and are comparable to or better than prior spatially-aware and non-spatially-aware methods.

{
\bibliography{refs}
\bibliographystyle{abbrv}
}

\end{document}